\documentclass[letterpaper, 10 pt, conference]{ieeeconf} 
\usepackage{cite}
\usepackage{graphicx}
\usepackage{float}
\usepackage{colortbl}

\IEEEoverridecommandlockouts                              

\title{\LARGE \bf
Magnetic Navigation of a Rotating Colloidal Swarm \\Using Ultrasound Images}
\author{Qianqian Wang$^{1}$, Lidong Yang$^{1}$, Jiangfan Yu$^{1}$, Chi-Ian Vong$^{1}$, Philip Wai Yan Chiu$^{2,3}$ and Li Zhang$^{1,2,4}$
\thanks{*The research work is financially support by the General Research Fund (GRF) with Project No. 14209514, 14203715 and 14218516 from the Research Grants Council (RGC) of Hong Kong SAR. }
\thanks{$^{1}$ Q. Wang, L. Yang, J. Yu, C. Vong and L. Zhang are with the Department of Mechanical and Automation Engineering of The Chinese University of Hong Kong (CUHK), Shatin NT, Hong Kong SAR, China.
        {\tt\small lizhang@mae.cuhk.edu.hk}}%
\thanks{$^{2}$ Chow Yuk Ho Technology Center for Innovative Medicine, The Chinese University of Hong Kong, Shatin NT, Hong Kong SAR, China. 
        }%
\thanks{$^{3}$ Department of Surgery, The Chinese University of Hong Kong, Shatin NT, Hong Kong SAR, China. 
        }%
\thanks{$^{4}$ CUHK T Stone Robotics Institute, The Chinese University of Hong Kong, Shatin NT, Hong Kong SAR, China. 
        }%
}

\begin{document}

\maketitle
\thispagestyle{empty}
\pagestyle{empty}

\begin{abstract}
Microrobots are considered as promising tools for biomedical applications. However, the imaging of them becomes challenges in order to be further applied on \emph{in vivo} environments. 
Here we report the magnetic navigation of a paramagnetic nanoparticle-based swarm using ultrasound images.
The swarm can be generated using simple rotating magnetic fields, resulting in a region containing particles with a high area density.
Ultrasound images of the swarm shows a periodic changing of imaging contrast. 
The reason for such dynamic contrast has been analyzed and experimental results are presented.
Moreover, this swarm exhibits enhanced ultrasound imaging in comparison to that formed by individual nanoparticles with a low area density, 
and the relationship between imaging contrast and area density is testified. 
Furthermore, the microrobotic swarm can be navigated near a solid surface at different velocities, and the imaging contrast show negligible changes.
This method allows us to localize and navigate a microrobotic swarm with enhanced ultrasound imaging indicating a promising approach for imaging of microrobots.
\end{abstract}

\section{Introduction}
Wirelessly actuated microrobots are able to perform effective motion in fluid environments, which can be applied for biomedical applications \cite{nelson2010microrobots,sitti2015biomedical}.
Among them, microrobots actuated by magnetic fields have been extensively studied \cite{dreyfus2005microscopic,zhang2009artificial,yang2018model} and emerge as promising tools for \emph{in vivo} applications.
To further explore applications of microrobots in a living body, \emph{in vivo} imaging of microrobots is an essential issue that needs to be addressed \cite{medina2017medical}. 
Taking advantage of medical imaging techniques, localization of microrobots has been investigated recently (e.g., magnetic resonance imaging (MRI) \cite{martel2009mri}, positron emission tomography (PET) \cite{vilela2018medical}, \emph{in vivo}  fluorescence imaging (IVIS) \cite{servant2015controlled}).
Among these imaging techniques, ultrasound imaging stands out as one of the most promising imaging tools for microrobots due to its relatively low cost, deep imaging depth (approximately 10 cm within the human body) and maturity of the technology. Using the feedback from ultrasound images, motion control and path planning of a millimeter-scale robotic gripper can be realized \cite{scheggi2017magnetic}, such feedback can be further used for guided rubbing of blood clots \cite{khalil2018mechanical}.

\begin{figure}[!t]\centering
	\includegraphics[width=0.95\columnwidth]
    {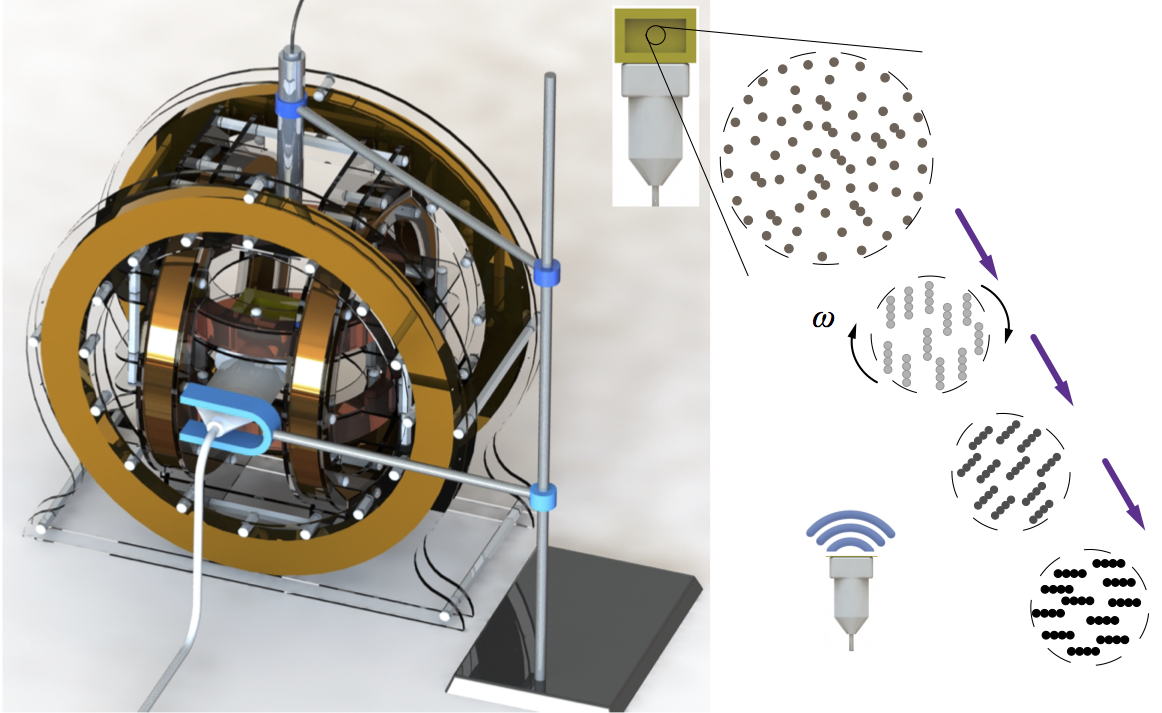}
    \caption{Schematic of the robotic swarm with dynamic ultrasound imaging contrast. Magnetic nanoparticles are filled into the tank and placed at the center of the Helmholtz coil system. The ultrasound images are acquired by an ultrasound system (Terason t3200, Teratech Corporation, USA). The diffused nanoparticles are gathered to form a swarm using a rotating magnetic field. A camera is mounted on the top for observation. The contrast of imaging is dependent on the angle between chain orientation and propagation direction of ultrasound waves. The black arrow shows the rotation direction of particle chains and blue arcs refers to the propagation of ultrasound wave.}
    \label{fig1}
\end{figure}

However, resolution of ultrasound imaging has several limitations and still remains challenges for imaging of microrobots. It relies on gradients of acoustic impedance and the scale of microrobot should be larger than the sonographic detection limit. 
To address this issue, one straightforward method is the usage of microrobot at a relatively large scale (e.g, millimeter-scale) \cite{scheggi2017magnetic,khalil2018mechanical,hu2018small}. However, this may encounter a critical limitation for further applications in confined environments. 
The second method is the usage of microbubbles.
Microbubbles are small (typically 1-8 $\mu m$ in diameter) gas-filled microspheres and they are used as ultrasound contrast agents in clinical imaging to improve the imaging quality due to increased scattering and reflection of ultrasound waves\cite{kiessling2011application}.
Although microbubbles have been successfully developed in clinical imaging \cite{cosgrove2009clinical}, 
several drawbacks limit their applications, such as low stability and short half-life \cite{min2012gas}. 
Microjet, as a kind of bubble-driven microrobots, exhibits propulsion through catalytic reaction with surrounding fluids. These microrobots can be tracked indirectly by the generated tails of bubbles using ultrasound images \cite{sanchez2014magnetic,olson2013toward}. However, this method requires specific surrounding environments for catalytic reaction, and the reaction rate and time are hard to control. 

A single microrobot at small scale challenges the imaging quality, making it hard for localization. 
To address this issue, swarm control and manipulation are worth to be investigated. 
The building blocks of a swarm are usually at a small scale, they can be injected into a confined environment (e.g. blood vessel), navigated to a planned location and regathered again. 
A swarm of microrobots may enhance the imaging contrast for medical imaging compared to a single microrobot, such as MRI \cite{yan2017multifunctional}, PET \cite{vilela2018medical} and IVIS system \cite{servant2015controlled}. 
Besides,  larger doses of drugs (cargoes) can also be delivered using a swarm. 

In this study, we propose the localization and navigation of a magnetic nanoparticles-based robotic swarm using ultrasound imaging and magnetic field. 
We use $\rm{Fe_3O_4}$ nanoparticles as building blocks of the magnetic swarm. 
In Section \uppercase\expandafter{\romannumeral2}, the generation of the swarm is modeled. The experimental methods and generation of the swarm are demonstrated in Section \uppercase\expandafter{\romannumeral3}. In Section \uppercase\expandafter{\romannumeral4}, the periodic changing of imaging contrast is analyzed and experimentally presented, which shows the contrast is related to the orientation of particle chains insides the swarm  (Fig. \ref{fig1}). The swarm exhibits enhanced imaging contrast due to the higher area density of particles ($\sim 5  \mu g /mm^2$), and the relationship between area density and imaging contrast is investigated. Moreover, in this section, we demonstrate that the swarm can be navigated in a controlled manner. Finally, Section \uppercase\expandafter{\romannumeral5} concludes and provides directions for future work. 

\section{Mathematical modeling}
\subsection{Nanoparticles under a static magnetic field}
The nanoparticles are treated as nanospheres with particle radius of $a$. When an external magnetic field of magnitude $B$ is applied, the induced dipole moment of a particle is
\begin{equation}
\mu=\frac{4}{3}\pi a^3 \mu_0\chi B
\end{equation}
where $\chi$ and $\mu_0$ are magnetic susceptibility and permeability of vacuum, respectively. The magnetic force between particles can be described as \cite{biswal2004rotational}
\begin{equation}
\mathbf F= \frac{3\mu^2}{4\pi\mu_0 r^4}(3\cos^2\alpha-1)\mathbf{\hat{r}} + \frac{3\mu^2}{4\pi\mu_0 r^4}\sin(2\alpha) \bf \hat{\theta}
\end{equation}
where $\alpha$ is the phase lag between the direction of the external field and the chain orientation.
\begin{figure}[!t]\centering
	\includegraphics[width=0.8\columnwidth]
    {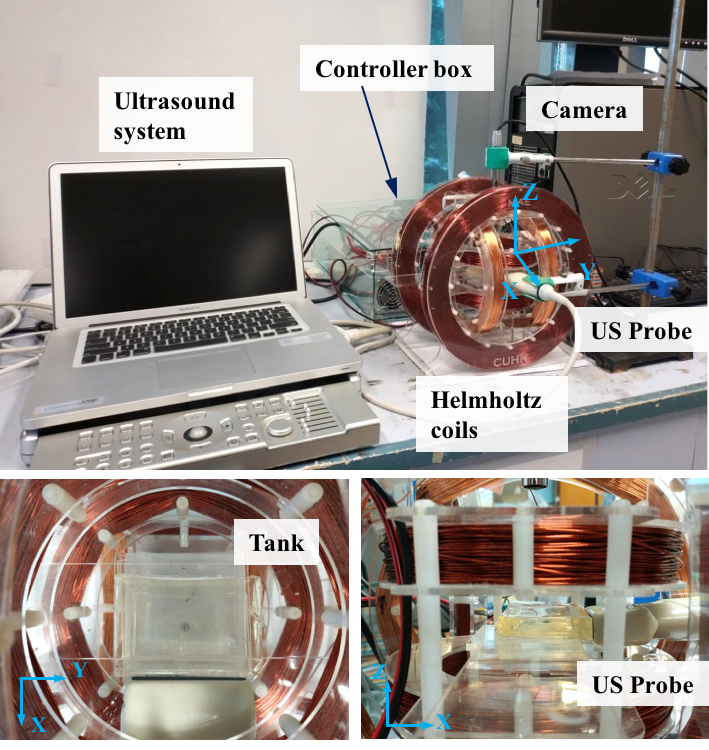}
    \caption{Experimental setup. The ultrasound probe (15L4, Teratech Corporation, USA) is placed at the center of the Helmholtz coils. The tank is placed at the center of the coil system with nanoparticles inside. Images are acquired by the ultrasound system (Terason t3200, Teratech Corporation, USA). The coils are controlled by a controller box and a lab PC. A miniature camera is mounted on the top of the coil for recording video. }
     \label{fig2}
    \end{figure}
\subsection{Nanoparticles under a rotating magnetic field}
The nanoparticles are gathered and then form particle chains due to dipole interaction. In order to describe nanoparticle chains actuated by a rotating magnetic field, we consider a chain of $N$ particles with a total length of $L=2Na$. In this paper, the Reynolds number is approximately $1\times 10^{-3}$, and the governing mechanism is the counterbalance between induced magnetic torque and drag torque of a particle chain due to the viscosity of the fluid. The induced magnetic torque of a rotating chain can be obtained as a sum of all the torques exerted by the neighboring particles. For simplicity, if only the magnetic interactions between nearest neighbors are taken into consideration, the magnetic torque can be expressed as \cite{Singh2005}
    \begin{figure*}[!t]\centering
 	\includegraphics[width=2\columnwidth]
     {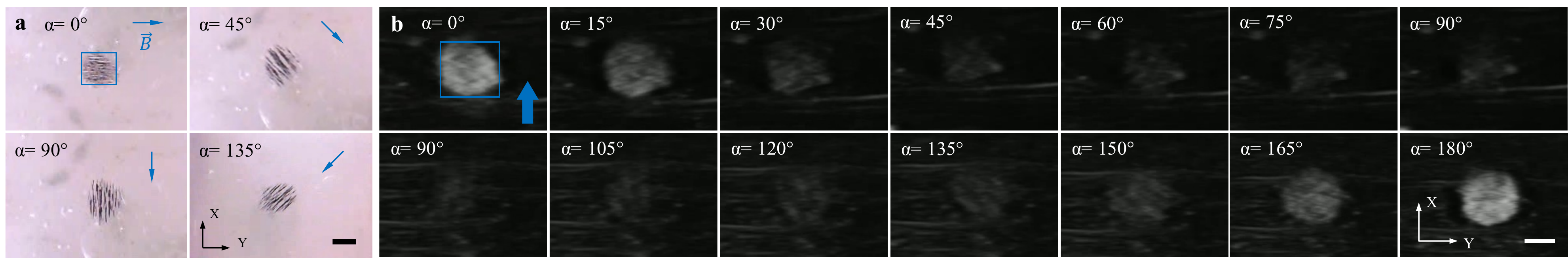}
    \caption{(a) Particle chains are formed by applying an in-plane static magnetic field at a strength of 8 mT. The orientation of the chains can be adjusted by changing the yaw angle ($\alpha$) of the field. Blue arrows refer to the direction of the external field. The scale bar is 2 mm. (b) Ultrasound images of nanoparticles under an in-plane static field. The yaw angles are changed from $0^{\circ}$ to $180^{\circ}$ with an interval of $15^{\circ}$. Blue rectangle refers to the region of interest (ROI). Arrow refers to the propagation direction of ultrasound waves. All the images are marked with the same ROI. The scale bar is 2 mm.}
    \label{fig3}
     \end{figure*}
\begin{equation}
\Gamma_m = \frac{3\mu^2}{4\pi\mu_0}\frac{N^2}{2(2a)^3}\sin(2\alpha)=\frac{\pi \mu_0  a^3 \chi^2 N^2}{12}B^2\sin(2\alpha)
\end{equation}
The viscous torque opposing the rotation of a chain with angular velocity $\omega$ can be calculated as \cite{doi1988theory}
\begin{equation}
\Gamma_d = \xi_r \omega = \kappa V \eta \omega
\end{equation}
where $\kappa$ is a shape factor, $V = N(4/3)\pi a^3$ is the volume of the chain and $\eta$ is the fluid viscosity. The shape factor of a linear chain consisting of $N$ particles, including hydrodynamic interactions is 
$\kappa = {2N^2}/{\ln(\frac{N}{2})}$. 
When a particle chain rotates in a steady angular velocity, the two torques are balanced at the chain center. 
Therefore, a defined phase lag can be obtained with a given magnetic chain strength and frequency, to be
\begin{equation}
\sin(2\alpha) = \frac{32N\eta\omega}{\mu_0\chi^2 B^2\ln(N/2)}
\end{equation}
If we have the condition $\sin(2\alpha)<1$, the particle chains rotate synchronously with the external field. In our experiments, all the particle chains exhibit synchronous rotation with the external rotating field. 

\section{Experimental Setup and Methods}
\subsection{System Description}
      \begin{figure}[!t]\centering
 	\includegraphics[width=1\columnwidth]
     {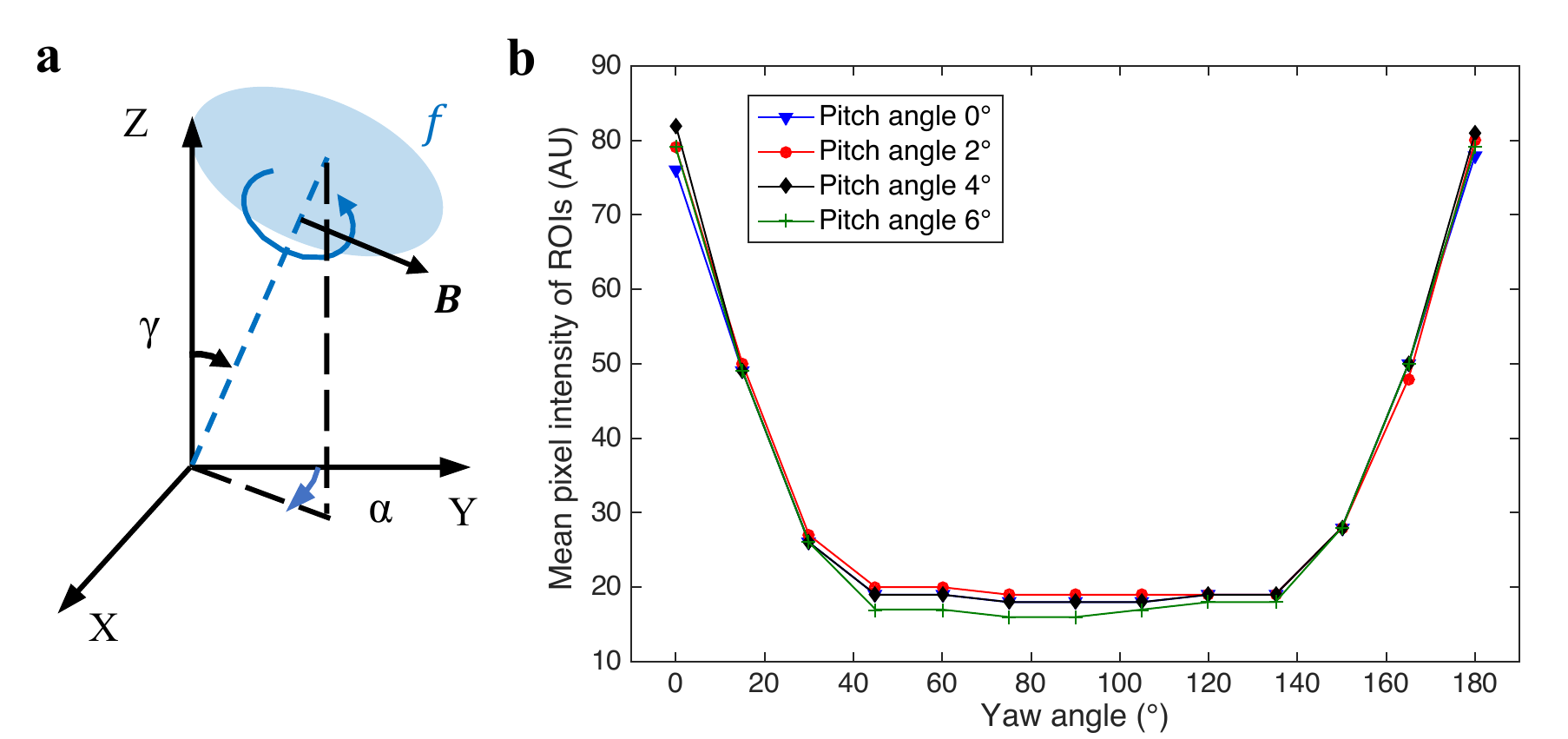}
    \caption{(a) Schematic of the rotating magnetic field. The blue line and arrow refer to the normal line and rotation direction of the magnetic field, respectively, $\alpha$ and $\gamma$ are the yaw angle and pitch angle. (b) Mean pixel intensity of ROIs when the nanoparticles under a static magnetic field with pitch angles of $0^{\circ}$, $2^{\circ}$, $4^{\circ}$ and $6^{\circ}$. The ROIs are shown in Fig. \ref{fig3} (b).}
       \label{fig4}
     \end{figure}
Magnetic actuation is achieved using a three-axis Helmholtz coil system. The generated rotating magnetic fields can be controlled by the control PC, and the relevant field parameters are controlled by an I/O card (Model 826, Sensoray Inc.) through the controller box (Fig. 2).
Air-free coupling between the ultrasound transducer and the swarm is achieved using gelatin. A gelatin tank (15 wt\%) with inner space of 45$\times$25$\times$5 mm is placed at the center of the coil system, and filled with 2 wt\% Polyvinylpyrrolidone (PVP) solution. The low echogenicity of gelatin provides a better imaging environment for our swarm.
The usage of PVP solution provides viscosity for stable generation of the swarm. 
An ultrasound system (Terason t3200, Teratech Corporation, USA) is integrated to the magnetic actuation system for imaging of the swarm. A linear array transducer (15L4, Teratech Corporation, USA) with bandwidth 15-4 MHz is mounted near the side wall of the gelatin with ultrasound gel, as shown in Fig. 2. 
The distance between the swarm and the transducer can be increased to localize the swarm at relatively deeper distances. However, the wavelength and frequency of the propagating ultrasound waves are inversely proportional. 
High-frequency ultrasound waves generate images with higher resolution, which can only be used for objects located at a superficial level. Low-frequency waves are more suitable for a swarm located at a deeper distance, because high-frequency ultrasound waves are easy to be attenuated as the depth increases. 
In our experiments, we used the B-mode (2D) to display the swarm with an imaging depth of 30 mm. The mechanical index and thermal index both are 0.6, and the 2D gain is 45.
 A miniature camera is mounted on the top of the coil system for video recording. During experiments, the recorded video from the ultrasound system and miniature camera are set to be synchronous.
  \begin{figure*}[!t]\centering
	\includegraphics[width=1.9\columnwidth]
    {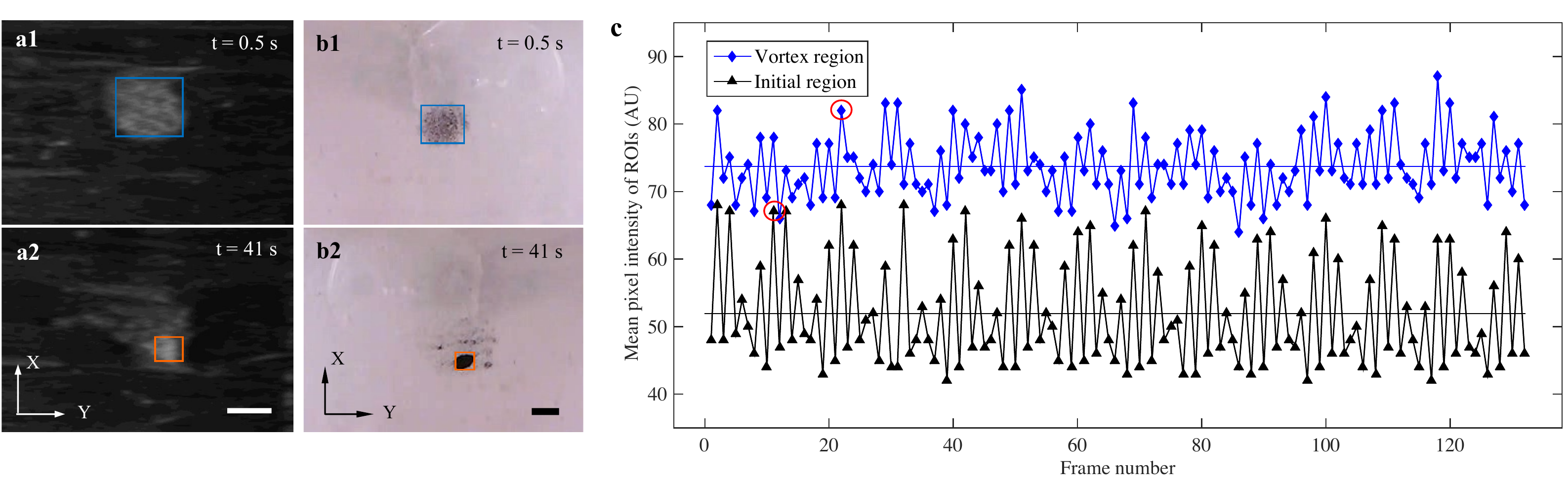}
    \caption{Ultrasound imaging contrast of the initial region and swarm region. (b1) and (b2) are acquired from the camera corresponded to (a1) and (a2), respectively. The initial region and swarm region are marked with blue and red rectangles in (a1,b1) and (a2,b2), respectively, as our ROIs. The Initial region refers to the region within 6 s after turning on the external field, and the swarm region includes a swarm that reaches equilibrium after 40 s. The scale bars are 2 mm. (c) The changing of mean pixel intensity of the two regions under a rotating magnetic field. The images of the ultrasound system are generated in 22 fps, therefore, the continuous 132 frames refer to 6 s (i.e. t = 0-6 s for the initial region and t = 40-46 s for the swarm region). Two straight lines are the mean values of intensity (i.e. 51.9 for the initial region and 73.7 for the swarm region), and two red circles respectively refer to the two frames of (a1) and (a2). The strength of the field is 8 mT, with an input frequency of 6 Hz.}
       \label{fig5}
    \end{figure*}
    \begin{figure}[!t]\centering
	\includegraphics[width=0.9\columnwidth]
    {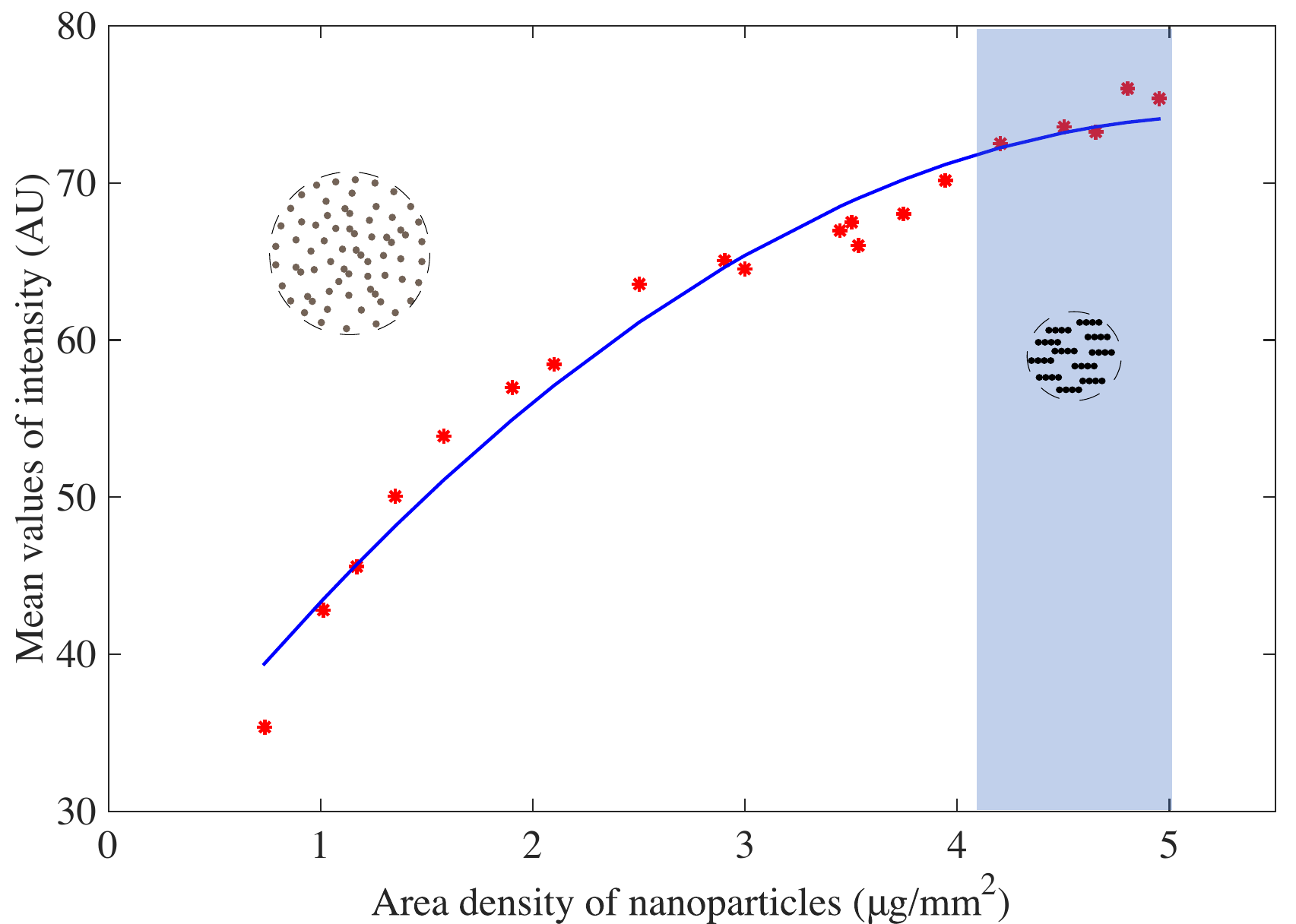}
    \caption{The mean values of intensity with different area density of nanoparticles. Each mean value of intensity is calculated from 66 continuous frames. The magnetic field strength and frequency are 8 mT and 6 Hz, receptively. Blue area refers to successfully generated swarm. The curves are fitting line of the data points. }
       \label{fig6}
    \end{figure}
\subsection{Generation of a particle-based swarm}
Here we use the magnetite paramagnetic nanoparticles ($\rm{Fe_3O_4}$) as the building blocks. Synthesis of the nanoparticles was previously reported \cite{deng2005monodisperse}. The average diameter of the particles is approximately 500 nm based on SEM images. 
One drop of nanoparticle suspension (2.5 $\mu L$, 9 $mg/mL$) is added to the gelatin tank using a pipette and the nanoparticles are diffused. A permanent magnet (with a surface magnetic field strength around 180 mT) is put under the tank and move slightly. Due to the magnetic field gradient, nanoparticles are gathered in a small area. Then the tank is put into the workspace of the Helmholtz coil for further actuation.
We use the previously reported method from our group to generate a swarm \cite{yu2017demand,ooo}.
After turning on the rotating field, rotating particle chains will be formed firstly due to the induced magnetic dipole-dipole interaction among particles. The interaction among particle chains lead to the gathering behaviors and a region with relative high-density particles can be observed. The region grows after attracting more particles and finally a particle-based swarm is generated at around t = 30 s. Some particle chains cannot be gathered which are far away from the swarm region, due to insufficient fluidic influence (see attached video).
\section{Experimental results and discussion}
\subsection{Nanoparticles under static magnetic field}
Since the swarm is formed by the gathering of nanoparticle chains, we firstly investigate the ultrasound images of nanoparticle chains.
By applying an in-plane static magnetic field with a strength of 8 mT,  particle chains are generated along the field direction (Fig. \ref{fig3} (a)). The orientation of the particle chains can be changed by changing the yaw angle ($\alpha$) of the external field due to the induced magnetic torque. 
Then the imaging contrast of nanoparticle chains is studied after the chains reach a static status. Ultrasound images are acquired with different yaw angles of the external field as shown in Fig. \ref{fig3} (b). The yaw angles are changed from $0^{\circ}$ to $180^{\circ}$ with an interval of $15^{\circ}$. The best contrast occurs when the yaw angles are $0^{\circ}$ and $180^{\circ}$, i.e. the orientation of particle chains is perpendicular to the propagation direction of ultrasound waves. The contrast decreases with increasing yaw angle from $0^{\circ}$ to $90^{\circ}$, while it increases with the yaw angle increasing from $90^{\circ}$ to $180^{\circ}$. 

In order to better understand the relationship between contrast and orientation of particle chains, the mean pixel intensity of region of interests (ROIs) is investigated quantitatively. The ROIs are defined as a region that includes all the nanoparticle chains as shown in Fig. \ref{fig3} (b), and the mean pixel intensity of ROIs are calculated using a LabVIEW program. 
The results are plotted in Fig. \ref{fig4} (b). The curves are approximately symmetry with respect to $90^{\circ}$, and the contrast reaches a minimal value when the yaw angle is around $90^{\circ}$. 
This is because when particle chains are perpendicular to the propagation direction of ultrasound waves (with yaw angles of $0^{\circ}$ and $180^{\circ}$), the scattered ultrasound waves reach the maximum value, resulting in the best contrast. While only little sound waves can be scattered  if the chains are parallel (with a yaw angle of $90^{\circ}$) with the propagation direction. 
In addition, by adding a small pitch angle ($\gamma$) to the external field, the particle chains are able to tilt from the substrate. Curves in Fig. \ref{fig4} (b) indicate that small pitch angles have a negligible influence on the imaging contrast, which means a small degree of tilt of chains will not change the amount of scattered sound waves significantly.  
\subsection{Nanoparticles under a rotating magnetic field}
In order to understand the imaging contrast of a rotating swarm, firstly we study the imaging contrast of nanoparticles before forming a swarm. 
By using a sequence of dynamic magnetic fields \cite{yu2017demand}, the nanoparticles are disassembled uniformly with a low area density ($\sim2  \mu g/mm^2$). 
After turning on a rotating magnetic field, the swarm cannot be formed in several seconds.
For better analyzing the process, an ROI named initial region is defined as shown in Fig. \ref{fig5} (a1), and the corresponding image obtained from the camera is shown in Fig. \ref{fig5} (b1). 
The 132 ultrasound images in Fig. \ref{fig5}(c) are continuously extracted from the recorded video of the ultrasound system (22 fps), which represents 0-6 s after turning on the external field. 
These frames are taken as our objects, and the changing of the mean pixel intensity of the initial region is obtained using a LabVIEW program. 
After 40 s, a swarm is successfully generated and an area with higher area density can be obtained ($\sim 4.5 \mu g /mm^2$). An ROI named swarm region is defined as shown in Fig. \ref{fig5} (a2). This region is defined as an inscribed rectangle of the gathered region of the particles. The mean pixel intensity of this ROI is obtained and results are shown in Fig. \ref{fig5} (c).  
Since the yaw angle is able to affect the imaging contrast, the two curves in Fig. \ref{fig5} (c) are in the same phase with respect to the yaw angle of the external field. 
Both curves show periodic changes because the swarm is not in a static status. As aforementioned, the swarm consists of numerous particle chains that are actuated by the external rotating field. Therefore, the angle between the orientation of chain and propagation direction of ultrasound wave changes, resulting in the periodic changes of contrast. 
Using this unique property, the swarm with dynamic contrast can be distinguished from other unnecessary objects or noise signal. 
The mean values of the two curves (i.e. 51.9 for the initial region and 73.7 for the swarm region) indicate that the swarm exhibits enhanced imaging contrast. 
This is because the swarm with high area density can scatter larger amount of ultrasound waves, comparing to nanoparticles in a low area density.

A quantitative study between area density and contrast is shown in Fig. \ref{fig6}. The nanoparticles with varied area density are realized by controlling the disassemble duration \cite{yu2017demand}. Then a rotating magnetic field at a frequency of 6 Hz is applied and imaging contrast is recorded. 
The mean pixel intensities are measured from continuously 66 frames (3s), and the mean values of intensity (y-axis) represent the mean of pixel intensity of the 66 frames, similar to the methods that used in Fig. \ref{fig5} (c).
The imaging intensity of the swarm is increased non-linearly with the area density of particles. 
The swarm have area densities larger than 4 $\mu g/mm^2$ and show an intensity higher than 70.
Interestingly, even nanoparticles in a very low area density ($< 1 \mu g /mm^2$) can also make a contribution to imaging contrast. 
During the formation of the swarm, some nanoparticles are not successfully assembled. These nanoparticles are assembled into short particle chains that rotate with the external field. From  Fig. \ref{fig5} (b2) we can observe that the region beyond the swarm also shows imaging contrast. Although the area density of this region is in a very low value, the ultrasound wave still can be scattered by nanoparticle chains.
 \begin{figure}[!t]\centering
	\includegraphics[width=0.9\columnwidth]
    {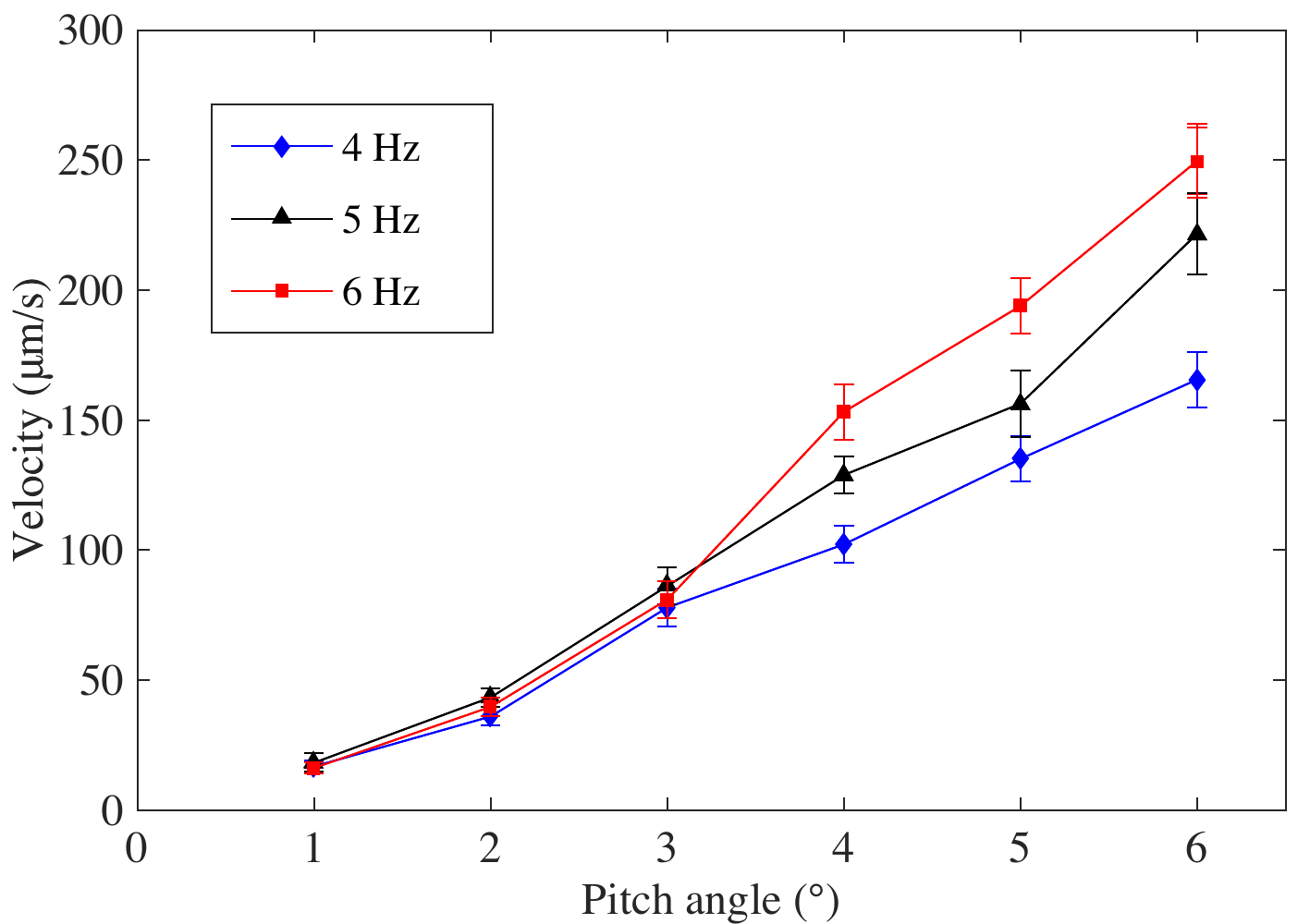}
    \caption{The relationship between velocity and pitch angle of the external fields with frequencies of 4 Hz, 5 Hz and 6 Hz. The magnetic field strength remains at 8 mT. Error bars are from data of three experiments.}
       \label{fig7}
    \end{figure} 
\subsection{2D navigation of the swarm}
\begin{figure}[!t]\centering
	\includegraphics[width=1\columnwidth]
    {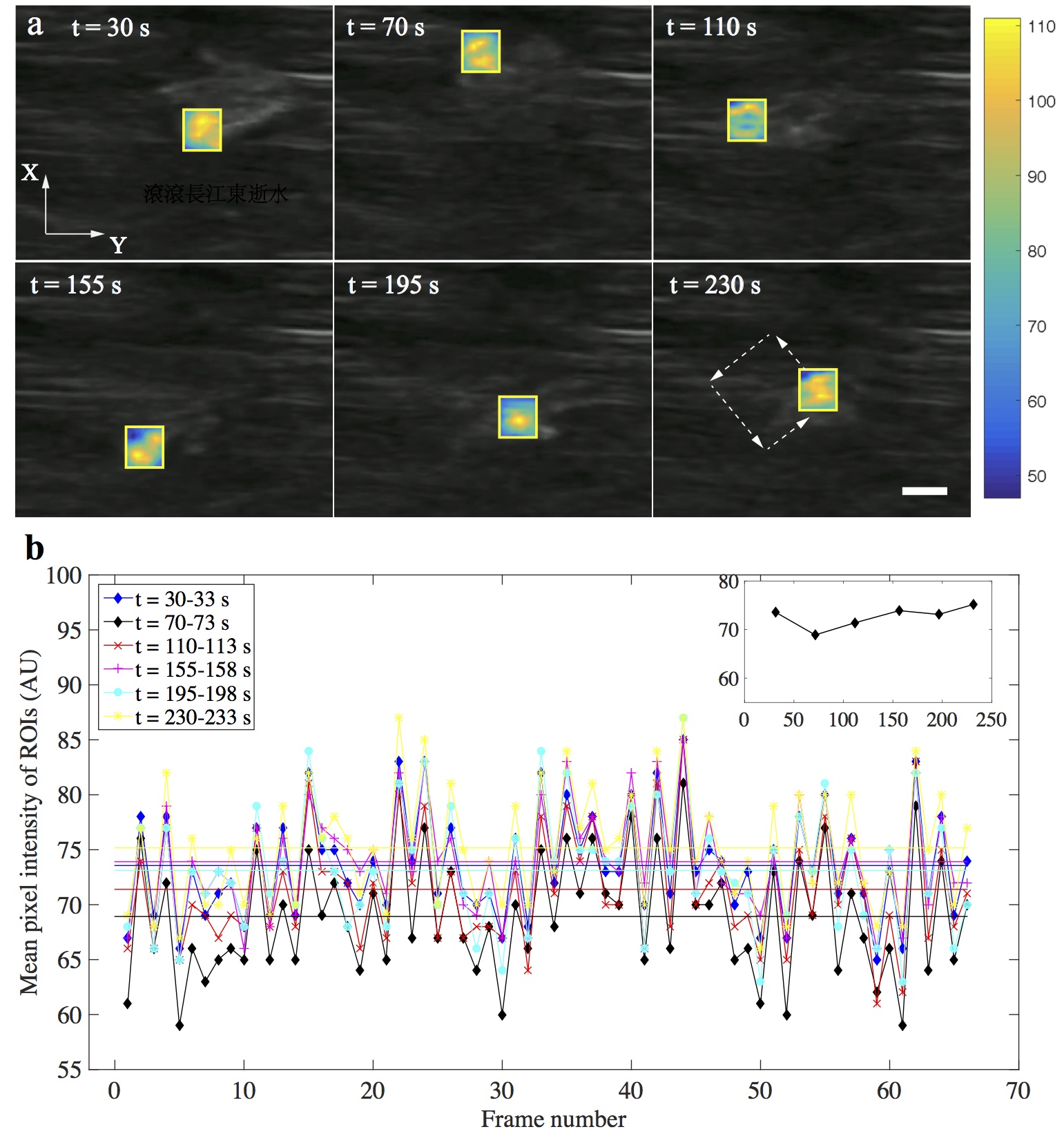}
    \caption{Magnetic navigation of the swarm. (a) A swarm is generated within 30 s and then navigated with a speed of 75 $\mu m/s$. The pixel intensity distribution of ROIs (yellow rectangle) are plotted by MATLAB, color bar refers to the value of pixel intensity. White arrows refer to the planned trajectory. The magnetic field has a strength of 8 mT and frequency of 6 Hz. The scale bar is 1 mm. (b) Mean pixel intensity of the ROIs during six navigation time slots. The continuous 66 frames refer to 3 s of the navigation process. Straight lines are the mean values of intensity corresponding to each curve. The inset is the relationship of the six mean values of intensity (Y-axis, unit: AU) against time (X-axis, unit: s).}
       \label{fig8}
    \end{figure}

Besides the usage of in-plane rotating fields, a swarm is able to exhibit locomotion by applying out-of-plane rotating field by adding a small pitch angle ($\gamma$). Pitch angle is a critical parameter for the motion of the swarm, and the existence of pitch angle induces a tilted angle between the swarm and the $XY$-plane. The locomotion is realized by friction asymmetry caused by the boundary (substrate) \cite{tierno2008controlled,sing2010controlled}.
The relationship between the pitch angles and translational velocities are plotted in Fig. \ref{fig7}. The swarm exhibits larger velocity when actuated by an external field with higher frequency. It is worth mentioning that during our experiments, the frequencies of the external field are below the step-out frequency. 
Motion direction of the swarm can be navigated in a controlled manner by adjusting the direction angle ($\alpha$). As shown in Fig. \ref{fig8} (a), the swarm is navigated as an entity to follow a rectangle trajectory with a speed of 75 $\mu m/s$ under ultrasound images (see attached video). 
Here we take the swarm region (yellow rectangle in Fig. \ref{fig8} (a)) as our ROIs. During the navigation, the pixel intensity distribution of the swarm region is plotted using a MATLAB program and the intensity data from ultrasound images. 
For a further understanding of the swarm during motion, we use the same frame processing method as that in Fig. \ref{fig5} (c). Here we take continuous 66 ultrasound images (3 s) as a time slot, and the changing of mean pixel intensity of the swarm region in the six time slot are plotted (Fig. \ref{fig8} (b)). All the curves have the same phase with respect to the yaw angle of the external field. 
The mean values of these intensity curves are compared in the inset of Fig. \ref{fig8} (b), showing that the contrast of the swarm is in a stable manner during navigation. 
The fluctuation of the mean value is caused by the disassembly and reassembly of nanoparticles in the swarm region since the particle-based swarm is a dynamic agent. 
All the particle chains exhibit two motion simultaneously, i.e. self-rotation and rotation around the center of the swarm. The two rotations are induced by the external field and equilibrium of the radial components of the interaction forces (i.e. fluidic, centrifugal and magnetic forces), respectively. 
Therefore, during the navigation the particle chains at the edge of the swarm undergo the greatest possibility of disassembly due to fluidic drag. We believe this is the main reason why the contrast shows a slight decrease at t = 70 s.
The disassembled nanoparticles exhibit the same motion direction with the swarm, and also can be reassembled into the swarm region if they are close enough to the swarm region. 
These experimental results show that the swarm can be navigated in a controlled manner, and the ultrasound images still have dynamic contrast with enhancement. This is essential for the usage of the swarm for real biomedical application. 
After being injected into the desired location, these nanoparticles can be regathered and a particle-based swarm can be generated. Due to the dynamic contrast, the swarm can be localized and distinguished from noise single.
Navigation of the swarm can be realized through the combination of ultrasound imaging and magnetic actuation. By using the regathered ability and the unique ultrasound imaging features, the swarm shows a great potential for \emph{in vivo} applications. 
\section{Conclusion} 
This work reports the localization and magnetic navigation of a nanoparticles-based swarm using ultrasound imaging and magnetic field. The swarm can be successfully generated and navigated using a simple rotating magnetic field. Experimental results show that the orientation of particle chains inside the swarm can significantly affect the imaging contrast, resulting in dynamic contrast properties. 
Furthermore, the robotic swarm is able to perform enhanced imaging contrast, compared to that formed by nanoparticles in a low area density. With small pitch angles, the swarm can be navigated on a solid surface with negligible changes of imaging contrast. 
The proposed method shows the potential for the \emph{in vivo} application of a microrobotic swarm using ultrasound images as the feedback. 
Further works in future will focus on real-time navigation under ultrasound guidance in different environments.
\section*{Acknowledgment}
We would like to thank Prof. Yongping Zheng from The Hong Kong Polytechnic University for fruitful discussion, and B. Wang from The Chinese University of Hong Kong for the synthesis of magnetic nanoparticles. 
\addtolength{\textheight}{-12cm}   



\bibliographystyle{./BIB/IEEEtran}
\bibliography{./BIB/IEEE_IROS}

\end{document}